%% file: paper.tex
\documentclass{article}

\usepackage[final]{corl_2024} 

\input{math_commands}

\usepackage{graphicx}
\usepackage{duckuments}
\usepackage{wrapfig}
\usepackage{booktabs}
\usepackage{siunitx}
\usepackage{dsfont}
\usepackage{amssymb}
\usepackage{subcaption}
\usepackage[capitalise]{cleveref}

\title{Analysing the Interplay of Vision and Touch\\for Dexterous Insertion Tasks}

\author{
  Janis Lenz$^{1}$
  \quad
  Theo Gruner$^{\dagger, 1,2}$
  \quad
  Daniel Palenicek$^{\dagger, 1,2}$
  \quad
  Tim Schneider$^{\dagger, 1}$
  \quad
  Jan Peters$^{1,2,3,4}$ \\
  $^\dagger$ equal supervision\quad
  $^1$ Intelligent Autonomuous Systems Group, TU Darmstadt\\
  $^2$ hessian.AI\quad
  $^3$ German Research Center for AI (DFKI)\quad
  $^4$ Robotics Institute Germany (RIG)
}

\begin{document}
\maketitle

\begin{abstract}
Robotic insertion tasks remain challenging due to uncertainties in perception and the need for precise control, particularly in unstructured environments. 
While humans seamlessly combine vision and touch for such tasks, effectively integrating these modalities in robotic systems is still an open problem. 
Our work presents an extensive analysis of the interplay between visual and tactile feedback during dexterous insertion tasks, showing that tactile sensing can greatly enhance success rates on challenging insertions with tight tolerances and varied hole orientations that vision alone cannot solve. 
These findings provide valuable insights for designing more effective multi-modal robotic control systems and highlight the critical role of tactile feedback in contact-rich manipulation tasks.
\end{abstract}

\keywords{dexterous insertion, tactile sensing, reinforcement learning}

\section{Introduction}
\begin{wrapfigure}[16]{r}{0.5\textwidth}
    \vspace{-1.2em}
    \includegraphics[width=0.5\textwidth]{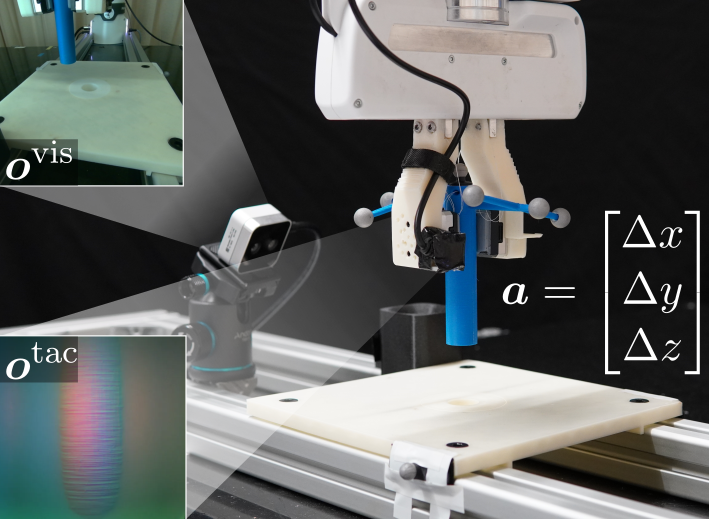}
    \caption{Dexterous insertion platform using vision and touch.
    $\vo^\text{vis}$ shows the visual observation and $\vo^\text{tac}$ the tactile observation.}
    \label{fig:setup}
\end{wrapfigure}
Humans solve insertion tasks, like plugging a plug into a socket, without much effort on a daily basis. 
Yet, in robotics insertion in an unstructured environment remains an open problem. 
The crucial challenge in insertion tasks or contact-rich manipulation tasks, in general, is uncertainty. 
The robot's perception system is prone to inaccuracy, so it usually does not know the plug's and socket's locations exactly. 
For successful insertion, however, precision is key as error margins are usually small.

When performing insertion tasks, humans rely on a combination of vision, touch, and compliance to achieve the necessary precision for successful insertion. 
With the recent rise of tactile sensors~\citep{biotac,yuan2017_gelsight,lambeta2020_digit,ward2018_tactip,funk2024_evetac}, both these sensing modalities are now available to robots. 
However, implementing a perception and control system that integrates both modalities and effectively deals with the remaining uncertainty remains challenging.

Robotic dexterous manipulation has been an active field of research for a long time. Many approaches rely on vision as their sole sensing modality~\citep{andrychowicz2020_inhandmanipulation,pinto2016_selfsupervisedgrasp,levine2018_handeye}. \citet{andrychowicz2020_inhandmanipulation} learn to rotate a cube in-hand into arbitrary target orientations using a combination of a vision-based pose estimation network and Proximal Policy Optimization~\citep{schulman2017proximal}. 
Instead of simulation,~\citet{pinto2016_selfsupervisedgrasp,levine2018_handeye} rely on large, autonomously collected datasets to learn grasping of various objects. 
While these works highlight the importance of vision for robotic manipulation tasks, vision alone is often insufficient, as it provides no information on forces acting at the contact points and suffers from occlusion by the robot's end-effector.

Prior work has utilized tactile sensing for dexterous manipulation~\citep{lach2023placing,Hogan,Kelestemur2022Mar,kim2022,dong2021_tactilerl}. 
Many recent approaches rely on vision-based tactile sensors ~\citep{Hogan,Kelestemur2022Mar,dong2021_tactilerl}, as they provide a rich, high-resolution representation of the contact patch while also allowing force estimation~\citep{funk2023highresolution}. 
A challenge is that tactile data is often difficult to interpret, making designing control loops with tactile sensors non-trivial. 
To address this, many prior works choose a data-driven approach to extract interpretable features from tactile data~\citep{Kelestemur2022Mar} or directly map tactile data to actions via end-to-end learning~\citep{lach2023placing,dong2021_tactilerl}.

Both tactile and vision sensors provide limited information about the environment. 
While tactile sensors can only provide information about objects the agent is in contact with, vision sensors suffer from occlusion and cannot measure contact forces. 
By combining both sources of information, one sensor can compensate for the other's shortcomings. 
The combination of vision and touch has been explored for 6D-pose estimation~\citep{dikhale2022visuotactile}, grasping~\citep{han2024_vistac_grasping,calandra2018_morethanafeeling}, object manipulation, and insertion~\citep{lee2020_makingsense,yu2024_mimictouch}. 
Like our work, \citet{lee2020_makingsense} also learns peg insertion with reinforcement learning from vision and touch. 
However, they first learn a representation of the multi-modal input with a Variational Auto Encoder (VAE)~\citep{kingma2014_vae} and then optimize the policy on the representation. 
While such a scheme can simplify the policy learning algorithm, it also means that the representation is not task-specific and potentially contains unnecessary information.

In this work, we explore the role of tactile sensing in improving robotic insertion tasks and analyze the interplay between visual and tactile inputs during complex manipulations.
We adapt the platform proposed in~\citep{palenicek2024learning} to include visual observations from an external camera and learn a policy using Dreamer-v3~\citep{hafner2023_dreamerv3}. 
Dreamer-v3 is an actor-critic algorithm that learns latent representations of sensory inputs to inform its policy.
Each input modality —- visual and tactile —- is concatenated and encoded into an embedding, which is fed into a recurrent state-space model that captures temporal dynamics essential for continuous control.
By jointly learning input representations and the policy, Dreamer-v3 effectively aligns the learning process of sensory inputs with action optimization.
The latent representation is learned via a VAE, enabling the model to generalize across complex manipulative tasks and adapt to variations in input signals.
This approach allows us to investigate how visual and tactile cues can complement each other in the context of precise and adaptive control.

\section{Autonomously Learning Visual-Tactile Peg Insertion in the Real World}
\label{sec:analysis}
\paragraph{Preliminaries.}
We consider solving a challenging insertion task from visual and tactile feedback, which will be formulated as an infinite horizon finite-time partially observable Markov decision process (POMDP). We denote observations by $\vo\in\mathcal{O}$, the (hidden) state by $\vh\in\mathcal{H}$, actions by $\va\in\mathcal{A}$, the reward by $r\in\sR$, and the discount factor by $\gamma\in[0,1]$. To solve the decision-making problem, we leverage \emph{Dreamer}~\citep{hafner2020_dreamer}, which learns a policy $\pi(\va_t| \vh_t)$ conditioned on the latent state representation. To deal with the partial observability, Dreamer learns a recurrent state-space model $p(\vh_t|\vh_{t-1}, \va_{t-1}, \vo_{t})$ \citep{hafner2019_planet}. 
To see the direct influence of the different input modalities on the action predictions, we rewrite the conditional dependence of the policy as $\pi(\va_t | \vh_{t-1}, \va_{t-1}, \vo_{t})$.
\paragraph{Hardware setup.}
Our setup is inspired by the work of~\cite{palenicek2024learning} and extends the authors' setup.
Figure~\ref{fig:setup} show our full task setup, which requires inserting a peg into a hole in the base plate.
The base plate is modular and the holes can be swapped for different tolerances $t$ with repect to the peg.
The peg is being held by a parallel gripper equipped with Gelsight sensors \citep{yuan2017_gelsight} at the finger tips. We add an external Intel RealSense camera \citep{realsense} for an external view of the scene. The observations are the downsampled RGB images of the scene camera $\vo^{\mathrm{vis}}\in\sR^{64\times64\times3}$ and the tactile sensor $\vo^{\mathrm{tac}}\in\sR^{64\times64\times3}$ at \SI{25}{\hertz}.
The policy outputs the relative new positions $\va = [\Delta x, \Delta y, \Delta z]^\intercal$ of the end-effector at \SI{10}{\hertz}.
The target positions are passed to the \emph{franky} control library~\cite{franky}, which integrates \emph{ruckig}~\cite{berscheid2021jerk} for smooth motion planning and leverages Franka's internal Cartesian impedance controller for execution. To ensure safe exploration, the workspace $\mathcal{W}$ is restricted to the area around the hole. The peg's pose is continuously tracked using OptiTrack~\citep{optitrack}, but this data is used exclusively for evaluation and not provided as input to the policy.
The reward function
\begin{equation*}
    \textstyle r = 
    \textstyle \underbrace{5\cdot (0.1 - |\vp_g - \vp_e|)}_{r_d:\:\text{proximity to the goal}} 
    \underbrace{+500\cdot \mathds{1}_{\{\mathcal{G}\}}(\vp_g)}_{r_g:\:\text{terminal reward upon reaching goal}\quad} 
    \underbrace{-50 \cdot \mathds{1}_{\{\mathcal{R}\}}(\vp_r)}_{r_r:\:\text{peg rotational penalty}\quad} 
    \underbrace{+\num{e-3} \cdot |\va|}_{r_a:\:\text{action penalty}}
\end{equation*}
is comprised of four components.
(i) $r_d$ proximity to the goal, 
(ii) $r_g$ a terminal reward upon reaching the goal $\mathcal{G} = \{\mathbf{x}\in\mathbb{R}^3: |\mathbf{p}_g - \mathbf{x}| < (5, 5, 5)[\si{\milli\metre}]\}$, 
(iii) $r_r$ a penalty to prohibit large rotational deviations $\mathcal{R} = \{\vtheta\,|\, |\vtheta - \vp_r| > 10^\circ\}$ of the end-effector rotation $\vp_r$, and 
(iv) $r_a$ an action penalty to encourage smooth motions. 
For more details of the base-setup, we refer to the previous work \citep{palenicek2024learning}.

\section{Experimental Results}
\label{sec:result}
\begin{figure}[t]
    \centering
    \begin{subfigure}[t]{0.32\textwidth}
        \centering
        \includegraphics[width=\textwidth]{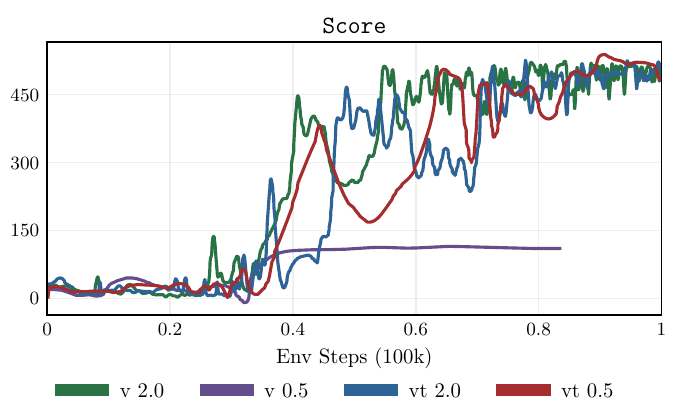}
    \end{subfigure}
    \begin{subfigure}[t]{0.32\textwidth}
        \centering
        \includegraphics[width=\textwidth]{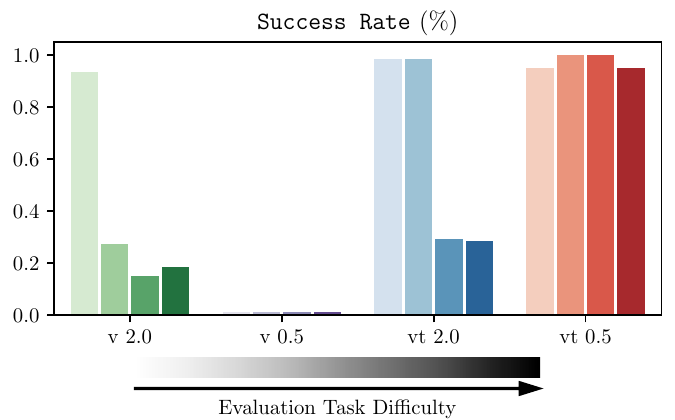}
    \end{subfigure}
    \begin{subfigure}[t]{0.32\textwidth}
        \centering
        \includegraphics[width=\textwidth]{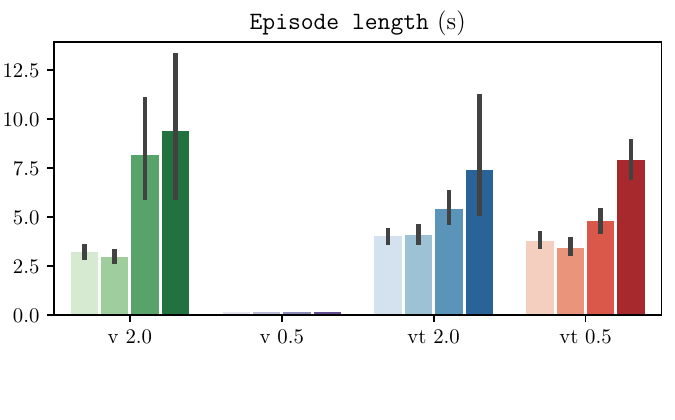}
    \end{subfigure}
    \vspace{1em}
    \caption{
    (left) Performances of the visual-only (v) and the visual-tactile (vt) policies trained on \SI{2}{\milli\metre} tolerance insertion hole and \SI{0.5}{\milli\metre} respectively.
    (middle) Final performance is reported as the success rate (middle) and the mean rollout length of successful insertions (right) over 20 trials across four varying insertion holes, with increasing evaluation task difficulty from left to right due to decreasing tolerances and increasing hole angles $(\mathrm{tol}, \alpha) \in \{(2, 0), (1, 4), (0.5, 0), (0.5, 4)\}[\si{\milli\metre}, ~^{\circ}]$.}
    \label{fig:performance}
\end{figure}

We present two studies, which are designed to showcase the importance of touch for dexterous insertion tasks.
The first involves training vision-based policies with and without the added sense of touch. The second analyzes the relative importance of each modality for action prediction.
\paragraph{The impact of tactile feedback during learning.}
We train four policies -- two based solely on vision and the other two combining vision with tactile feedback -- for insertion tasks involving different tolerances $t$ between the insertion hole and the peg.
The first set of experiments is performed for $\mathrm{tol} = \SI{2}{\milli\metre}$ over three seeds, where minimal jamming of the peg is expected.
The second configuration uses a hole with $\mathrm{tol} = \SI{0.5}{\milli\metre}$, which is anticipated to result in considerable jamming during insertion and increased interaction forces between the gripper and the peg. 
We report the training curves and the final performance of the respective models solving separate evaluation tasks in \Figref{fig:performance}.
In the training phase, all policies, except the vision-only policy trained on the \SI{0.5}{\milli\metre} tolerance, successfully ($r>500$) completed the peg insertion after 100k environment steps. 
During the evaluation, the \SI{2}{\milli\metre} and \SI{1}{\milli\metre} configurations could be solved reliably by all policies within 2.5 to 4~\si{\second}.
In the more challenging \SI{0.5}{\milli\metre} tolerance tasks, both the vision-only and vision-tactile policies exhibit a significant drop in success rates, except for the vision-tactile policy trained on the \SI{0.5}{\milli\metre} configuration, which maintains strong performance.
The increased episode length for successful insertions is directly related to more frequent jamming and tilting, highlighting the \texttt{vt 0.5} policy's ability to effectively handle these challenges.

\begin{figure}[t]
    \centering
    \includegraphics[width=\textwidth]{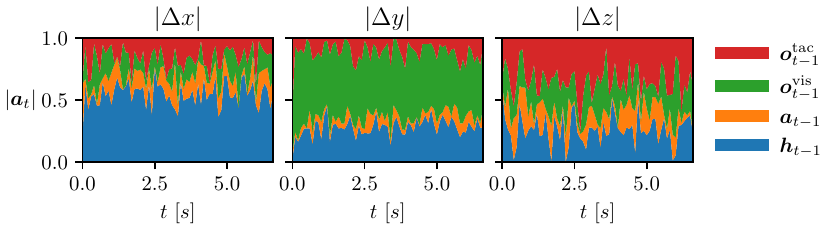}
    \caption{Shapley value analysis over a single exemplary trajectory of the \texttt{vt 0.5} experiment on the most challenging hole $(\mathrm{tol}=\SI{0.5}{\milli\metre}, \alpha=4^\circ)$. We report the individual contributions of the four input modalities on the action prediction in x-, y-, and z-direction.}
    \label{fig:shapley}
\end{figure}

\paragraph{Explaining the role of touch during an episode.}
To evaluate the influence of the two different observation modalities, we leverage Shapley value \citep{shapley1953_shapley} analysis. Shapley values provide a systematic way to assess the influence of different input components of $\vx
$ on the output of a model $f(\vx)$. In Dreamer, the input $\vx$ for the action prediction $\va_t = f(\vx)$ consists of the previous hidden state $\vh_{t-1}$, the visual $\vo^{\mathrm{vis}}_{t-1}$ and tactile information $\vo^{\mathrm{tac}}_{t-1}$, and the previous action $\va_{t-1}$. To evaluate the individual impact of each feature, we mask them by substituting their values with a default placeholder. This results in a total of $2^N, N=4$ distinct features, each identified by an integer label, allowing us to systematically analyze their respective contributions to the model's performance. Formally, we denote $\mathcal{F}$ as the set of the $N$ different features, then the Shapley value of feature $i\in\mathcal{F}$ is the attribution that the feature has on the outcome of the model
\begin{equation}
    \textstyle
    \phi_i(\vx) = |\mathcal{F}|^{-1}\sum_{\mathcal{S}\subset\mathcal{F}\setminus\{i\}}\binom{|\mathcal{F}| - 1}{|\mathcal{S}|}\left(f(\vx_{\mathcal{S}\cup \{i\}} - f(\vx))\right).
\end{equation}
Calculating Shapley values often becomes intractable due to the exponential scaling with $2^N$ \citep{lundberg2017_kernelshap}. However, in our case, with only 16 possibilities, we compute the Shapley values of the individual contributions exactly shown in \Figref{fig:shapley}. Along the x-axis, the model predominantly relies on its hidden state, with minimal influence from vision and tactile inputs. This may be attributed to the camera's orientation, which is primarily aligned along the x-direction, making depth perception more challenging. In contrast, along the y-axis, visual observations contribute most to action prediction, suggesting that visual feedback plays a key role in aligning the peg with the insertion hole. Finally, along the z-axis, tactile feedback has the greatest influence on action prediction, likely because vertical movements of the end-effector lead to increased tilts and jams, resulting in the highest impact on the contact forces applied between the peg and the hole.

\section{Conclusion}
\label{sec:conclusion}

In this work, we presented a comprehensive analysis of the interplay between visual and tactile feedback for robotic insertion tasks. 
Our results demonstrate that incorporating tactile sensing can significantly improve success rates on challenging insertions with tight tolerances (0.5mm) and varied orientations (up to 4° tilted insertion hole) that vision alone struggles to solve. 
Through Shapley value analysis, we revealed that different input modalities dominate action prediction along different axes -- vision primarily guides alignment in the camera plane, while tactile feedback is crucial for controlling vertical movements where contact forces are highest. 
However, it is essential to acknowledge that this analysis is confined to single trajectories and could benefit from additional explainability studies, such as gradient-based analyses. Furthermore, categorizing trajectories into distinct insertion stages may enhance the study's depth and facilitate statistical evaluation.

Our results suggest that the information the agent extracts from vision is effectively complemented by tactile sensor data.
In particular, when the hole tolerances are smaller and vision alone is insufficient for robust insertion, the agent learns autonomously to rely more heavily on tactile sensing.
Hence, our work demonstrates that RL is a powerful tool for learning robust policies from multi-modal sensor inputs.

As the vision-tactile policies manage to insert the peg successfully even for tight tolerances and varying inclination angles, increasing the task difficulty to different objects is a logical next step.
In particular, screw and light-bulb insertion is a challenging task with many real-world applications, which might significantly benefit from tactile feedback.
A crucial future direction is optimizing the representation learning part during policy optimization. To reduce the number of real-world interactions required, leveraging pre-trained vision and visual-tactile models could be beneficial.

\clearpage
\acknowledgments{This work was supported by the German Federal Ministry of Education and Research (BMBF) through the project Aristotle (ANR-21-FAI1-0009-01) and Hessian.AI.}

\bibliography{bibliography}  

\end{document}

%% file: math_commands.tex
\usepackage{amsmath,amsfonts,bm}

\def\Figref#1{Figure~\ref{#1}}

\def\eqref#1{equation~\ref{#1}}

\def\1{\bm{1}}

\def\vtheta{{\bm{\theta}}}
\def\va{{\bm{a}}}

\def\vh{{\bm{h}}}

\def\vo{{\bm{o}}}
\def\vp{{\bm{p}}}

\def\vx{{\bm{x}}}

\DeclareMathAlphabet{\mathsfit}{\encodingdefault}{\sfdefault}{m}{sl}
\SetMathAlphabet{\mathsfit}{bold}{\encodingdefault}{\sfdefault}{bx}{n}

\def\sR{{\mathbb{R}}}

